\documentclass[10pt,twocolumn,letterpaper]{article}

\usepackage{cvpr}              
\usepackage{tabularray}

\definecolor{cvprblue}{rgb}{0.21,0.49,0.74}
\usepackage[pagebackref,breaklinks,colorlinks,allcolors=cvprblue]{hyperref}

\def\paperID{} 
\def\confName{}
\def\confYear{}

\title{MagicQuill V2: Precise and Interactive Image Editing with Layered Visual Cues}

\author{Zichen Liu$^{\heartsuit,1,2}$, Yue Yu$^{\heartsuit,1,2}$, Hao Ouyang$^{2}$, Qiuyu Wang$^{2}$, Shuailei Ma$^{3,2}$, Ka Leong Cheng$^{2}$, Wen Wang$^{4,2}$, \\  
Qingyan Bai$^{1,2}$, Yuxuan Zhang$^{5}$, Yanhong Zeng$^{2}$, Yixuan Li$^{2,5}$, Xing Zhu$^{2}$, Yujun Shen$^{\dagger,2}$, Qifeng Chen$^{\dagger,1}$ \\[0.2cm]
$^1$HKUST, $^2$Ant Group, $^3$NEU, $^4$ZJU, $^5$CUHK
}
\begin{document}
\twocolumn[{
\renewcommand\twocolumn[1][]{#1}
\maketitle
\begin{center}
    \vspace{-10pt}
    \includegraphics[width=0.85\linewidth]{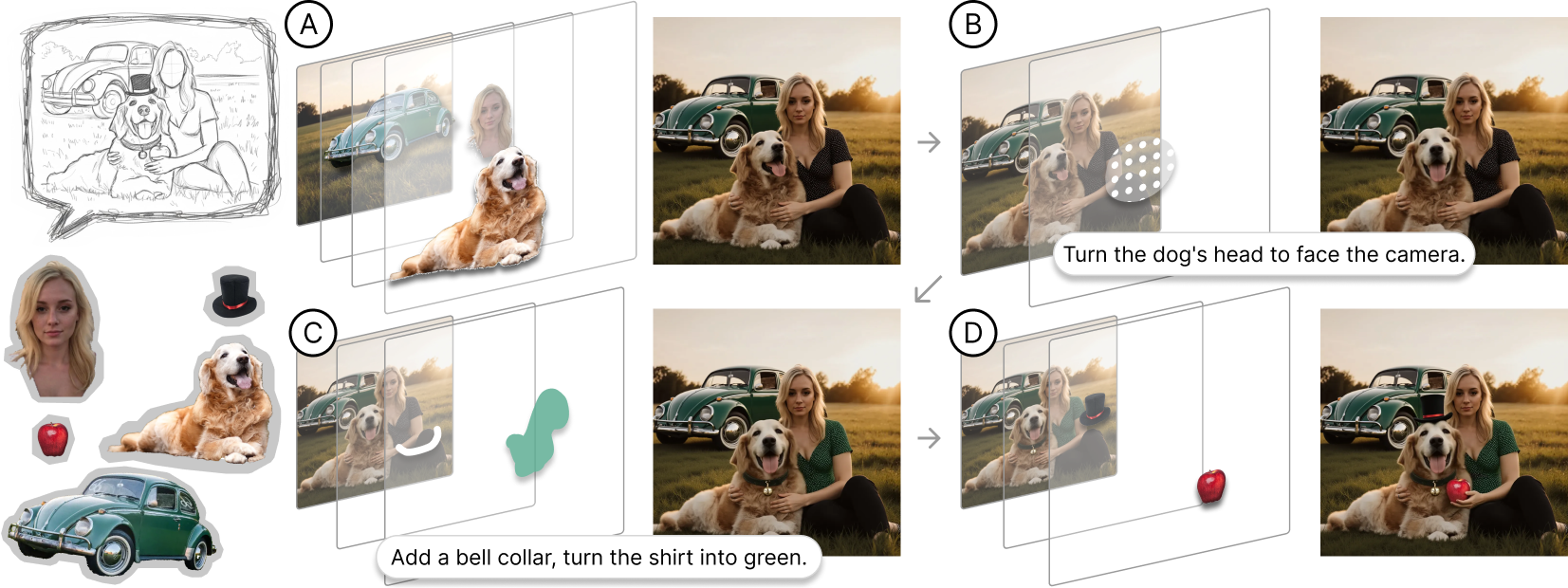}
    \vspace{-6pt}
    \captionsetup{type=figure}
    \caption{MagicQuill V2 introduces a layered composition framework for precise generative image editing.
    Users articulate complex intents by stacking independent visual layers in a continuous workflow: (A) composing a base scene with multiple \textbf{content layers} (car, lady, dog); (B) restructuring the dog's pose with a \textbf{spatial layer}; (C) modifying visual attributes by sketching a bell collar on the \textbf{structural layer} and painting the shirt on the \textbf{color layer}; (D) inserting final details like a hat and apple seamlessly. Demo available at \href{https://magicquill.art/v2/}{project page}.}
    \label{fig:teaser}
\end{center}
}]

\maketitle
\let\thefootnote\relax\footnotetext{\noindent$^\heartsuit$Equal contribution. $^\dagger$Corresponding author.}
\begin{abstract}
We propose MagicQuill V2, a novel system that introduces a \textbf{layered composition} paradigm to generative image editing, bridging the gap between the semantic power of diffusion models and the granular control of traditional graphics software. While diffusion transformers excel at holistic generation, their use of singular, monolithic prompts fails to disentangle distinct user intentions for content, position, and appearance. To overcome this, our method deconstructs creative intent into a stack of controllable visual cues: a content layer for what to create, a spatial layer for where to place it, a structural layer for how it is shaped, and a color layer for its palette. Our technical contributions include a specialized data generation pipeline for context-aware content integration, a unified control module to process all visual cues, and a fine-tuned spatial branch for precise local editing, including object removal. Extensive experiments validate that this layered approach effectively resolves the user intention gap, granting creators direct, intuitive control over the generative process.
\end{abstract}    
\section{Introduction}
\label{sec:intro}
The advent of diffusion transformers has ushered in a new era of powerful image editing systems, such as the FLUX series \cite{batifol2025kontext}, Qwen-Image \cite{wu2025qwen-image}, GPT-4o \cite{openai2024gpt4ocard}, and Nano Banana (Gemini 2.5 Flash Image) \cite{nano_banana}. These models exhibit remarkable proficiency in executing complex edits based on natural language or reference images. 

However, this reliance on high-level, holistic inputs presents a considerable barrier when users require precise control over the fundamental components of an image. 
As illustrated in~\cref{fig:teaser}, consider the nuanced task of creating a complex scene: a user starts by composing foundational elements like a vintage car, a person, and a dog (A). Then, the user may want to precisely adjust specific details, such as turning the dog’s head to face the camera (B), modifying attributes like shirt color (C) and inserting fine details like a top hat or an apple (D).
Existing conversation-based systems struggle to disentangle these intertwined commands from a single prompt, as they often sacrifice the spatial precision in traditional tools (e.g., Photoshop, GIMP).
After all, a user's creative intent is rarely monolithic; it is a composite of distinct desires concerning \textbf{\textit{what}} to create, \textbf{\textit{where}} to place it, \textbf{\textit{how}} it should be structured, and with \textbf{\textit{what colors}} it should be painted.

In this work, we introduce a novel framework built upon the principle of \textbf{layered composition}, drawing from the robust and intuitive workflows of professional graphics software. Our approach deconstructs the editing process by mapping each fundamental component of visual creation to its own dedicated, controllable layer. This allows a user to articulate their intent not as an ambiguous command, but as a stack of precise \textbf{visual cues}: a \textbf{content layer} specified by a foreground reference image to control what appears; a \textbf{spatial layer} defined by a mask to control where the editing appears; a \textbf{structural layer} guided by an edge map to dictate how it is shaped; and a \textbf{color layer} applied with strokes to determine its palette. Within our interactive system, these layers are not static inputs but dynamic elements that can be independently composed, repositioned, and refined. By uniting the generative prowess of diffusion models with the granular control of a layered interface, our approach offers a powerful yet intuitive solution that bridges the gap between semantic convenience and spatial precision.

We build our system upon the strong image editing model FLUX Kontext \cite{batifol2025kontext}. To effectively implement our layered design, we introduce several key contributions. First, we propose a specialized data generation pipeline to train the content layer (foreground pieces) for seamless, context-aware content integration, rather than a simple copy-paste operation. Second, we introduce a unified control module designed to process all control cues, including the spatial layer (mask), the structural layer (edge maps), and the color layer (color maps). The control module precisely constrains the geometry and palette accordingly. We additionally fine-tune the spatial layer branch on object removal data, enabling a dedicated and precise removal function. Finally, we design and implement a cohesive, interactive system that enables users to manage and compose these distinct layers intuitively. This system, augmented with tools like SAM \cite{kirillov2023sam} for effortless cue creation, realizes a precise, layer-like editing workflow that provides users with fine-grained control and creative freedom.

We demonstrate through extensive experiments that our method (MagicQuill V2) significantly outperforms existing approaches in editing that demands high precision. Our work validates that this layered composition approach effectively bridges the intention gap, granting users direct and powerful control over the generative process.

\section{Related Works}
\label{sec:rw}
\subsection{Controllable Image Generation and Editing}
While initial text prompts offered creative scope, their inherent ambiguity limited spatial precision. Early pioneering works like ControlNet \cite{zhang2023controlnet} and T2I-Adapter \cite{mou2024t2i-adapter} augmented UNet-based \cite{ronneberger2015unet, rombach2022SD} architectures to accept spatial conditions such as Canny edges, depth maps, and human pose, marking a significant step towards granular control.
With the architectural shift towards more powerful Diffusion Transformers \cite{peebles2023dit}, new control paradigms emerged~\cite{tan2024ominicontrol,tan2025ominicontrol2,zhang2025easycontrol,cai2025diffusion,cao2025relactrl,mao2025ace++,jiang2025vace}. 
OminiControl \cite{tan2024ominicontrol, tan2025ominicontrol2} introduced a parameter-efficient, unified framework by concatenating text, noise, and condition tokens into a single sequence for joint processing. 
In contrast, EasyControl \cite{zhang2025easycontrol} proposed a modular approach, processing each condition in an isolated branch to enhance plug-and-play compatibility.

Despite their architectural advance, these state-of-the-art methods, and even more recent multi-modal models that accept multiple reference images like Qwen-Image \cite{wu2025qwen-image}, Step1X-Edit~\cite{liu2025step1x}, GPT-4o \cite{openai2024gpt4ocard}, Nano-Banana \cite{gemini2.5}, and Seedream \cite{seedream2025seedream4}, share a fundamental limitation. 
Even when a reference image is provided, its impact on the editing process is governed by the text prompt. This paradigm inherits the natural imprecision of language, as text prompts are inherently inefficient for specifying the exact where, what, and how of a targeted spatial edit.
MagicQuill V2 challenges this paradigm by introducing precise visual cues, such as foreground pieces or a sketched edge map, that serve as a direct complement to the textual prompt.
These cues provide the specific spatial and appearance information that text cannot, eliminating the inferential gap and creating a direct channel between user intent and generative output.

\subsection{Layered Image Synthesis}
Layered image synthesis encompasses both layer decomposition \cite{quattrini2024alfie, fontanellagenerating, yang2024generative, Text2Layer, kang2025layeringdiff, yang_Layer_Decomposition, tudosiu2024mulan, huang2024layerdiff, dalva2024layerfusion, PSDiffusion} (analyzing an image into layers) and layer composition (assembling generative elements), which our work aligns with. 
This contrasts with the related subtask of image harmonization \cite{guo2021image_harmonization,pitie2005n,cohen2006color,tao2013error,sunkavalli2010multi,zhu2015learning,tsai2017deep,dovenetharm,jiang2021ssh,Ling2021region,Hao2020ImageHW,cong2020bargainnet,Sofiiuk2021fore,chen2024deep,FreeCompose,qi2018semi}, which assumes a pre-existing, static foreground to be adjusted to the scene. In our setting, the ``foreground'' is not a fixed layer but a generative cue that guides what the model should synthesize, allowing for flexible manipulation and interaction with the context through generative instructions.

Image composition is a broader task that involves inserting objects into a scene while preserving their identity \cite{niu2021survey_image_composition}. 
Representative works include Paint-by-Example~\cite{yang2023paint_by_example}, TF-ICON~\cite{lu2023tficon}, AnyDoor~\cite{chen2024anydoor}, MimicBrush~\cite{chen2024zero}, Imprint~\cite{song2024imprint}, Insert-Anything~\cite{song2025insert_anything}, UniReal~\cite{chen2025unireal} and IC-Custom~\cite{li2025ic_custom}.
Although these methods allow users to specify a layout or mask to control where an object should appear, the final appearance is still inferred from a holistic reference image, which constrains fine-grained control. 
Our approach differs fundamentally: by enabling users to provide pieces of the foreground as direct cues, we offer a more precise channel to guide the editing process.

\subsection{Interactive Support for Expressing User Intent}
A significant challenge in generative editing is bridging the ``intention gap'', the discrepancy between a user's precise mental image and the ambiguous modalities available to describe it.
While early solutions relied on autonomously suggesting or refining prompts via Multimodal Large Language Models (MLLMs)~\cite{fu2023mgie, liu2025magicquill}, recent Human-AI interaction research has shifted toward more direct, expressive paradigms that allow users to express intent through interactive manipulation rather than just description.

To overcome the linearity of chat-based prompting, some approaches propose tools like Fillable Brushes to reify abstract user intents into direct and interactive graphical ``instruments.''~\cite{riche2025ai}
Some other approaches, like SketchFlex~\cite{lin2025sketchflex}, focus on addressing the barrier of artistic expertise, leveraging MLLMs not just for interpretation, but to actively refine novice users' rough sketches into coherent spatial conditions.
Most relevant to our work are approaches that decouple visual attributes for more granular control. FusAIn~\cite{peng2025fusain} introduces the concept of ``smart pens'' that allow designers to explicitly extract and apply specific visual attributes such as object identity, color, or texture from inspiration images.

MagicQuill V2 formalizes these interactive concepts into a layered editing framework for professional precision. Instead of interpreting rough sketches or using abstract instruments, we map fundamental visual cues to explicit layers, allowing users to dynamically compose and refine the entire generative process in real-time.


\section{Methodology}
\label{sec:method}
Built upon FLUX Kontext \cite{batifol2025kontext}, our methodology is designed to realize the layered composition framework introduced previously. Our goal is to develop a system that generates a target image $x$ conditioned jointly on a context image $y$, a natural language instruction $c$, and a stack of precise layered visual cues $L$. These cues are mapped directly to the fundamental components of visual creation. Formally, we aim to approximate the conditional distribution:
\begin{equation}
p(x | y, c, \{L_{fg}, L_{control}\}).
\end{equation}

Here, the conditioning cues are explicitly divided to match our layered framework: The \textbf{content layer} ($L_{fg}$), defining what to create, is specified by one or more foreground pieces ${F_i}$. The \textbf{control layers} ($L_{control}$) provide explicit control defining where, how, and with what colors to edit. This group consists of \textbf{spatial layer} (mask $M$) for targeted regional editing, \textbf{structural layer} (edge map $E$) for precise geometric guidance, and \textbf{color layer} (color map $C$) for exact color control.

This section details the construction of MagicQuill V2. First, we introduce our data generation pipeline to enable the content layer (\cref{subsec:content_layer}). Second, we detail our unified control architecture for the control layers (\cref{subsec:control_layer}). Finally, we present the interactive system that unifies these layers into an intuitive, precise editing tool (\cref{subsec:UI}).

\subsection{Enable Content Layer via Foreground Cues}
\label{subsec:content_layer}
The content layer ($L_{fg}$), which defines what to create, consists of one or more user-provided foreground pieces ($F_i$) that function as content anchors and an optional mask to specify editing regions. The core task is to execute the edit by seamlessly integrating the content and generating the surrounding context, rather than a simple copy-paste operation. We enable this capability through a novel data generation pipeline.

\begin{figure}[t]
    \centering
    \includegraphics[width=\linewidth]{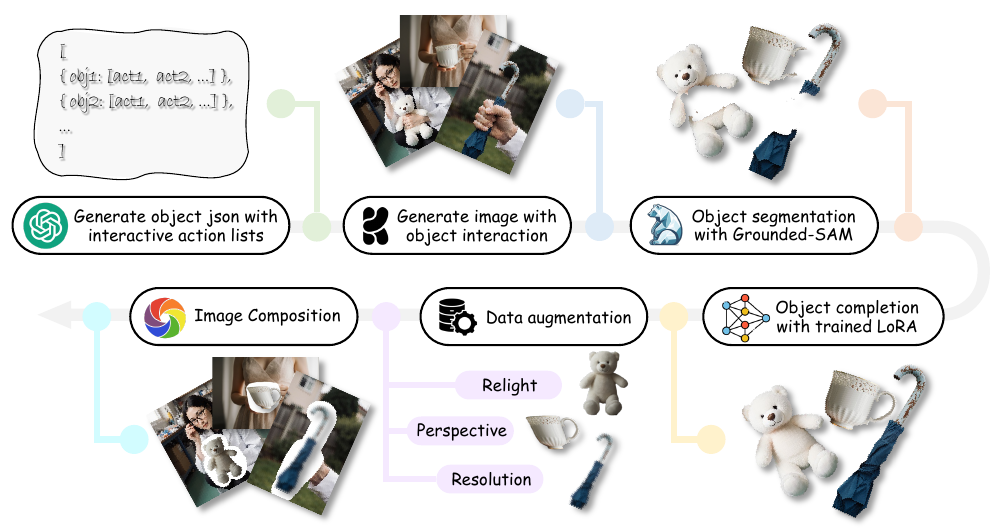}
    \caption{Overview of our Data Construction Pipeline. We start by synthesizing images depicting interactions. From these, we extract foreground objects, restore occluded items using a trained object completion LoRA, apply a series of augmentations (Relight, Perspective, Resolution), and finally composite the augmented object back into the scene to create the training data.}
    \vspace{-0.4cm}
    \label{fig:data_piepline}
\end{figure}

\begin{figure*}[t]
    \centering
    \includegraphics[width=\linewidth]{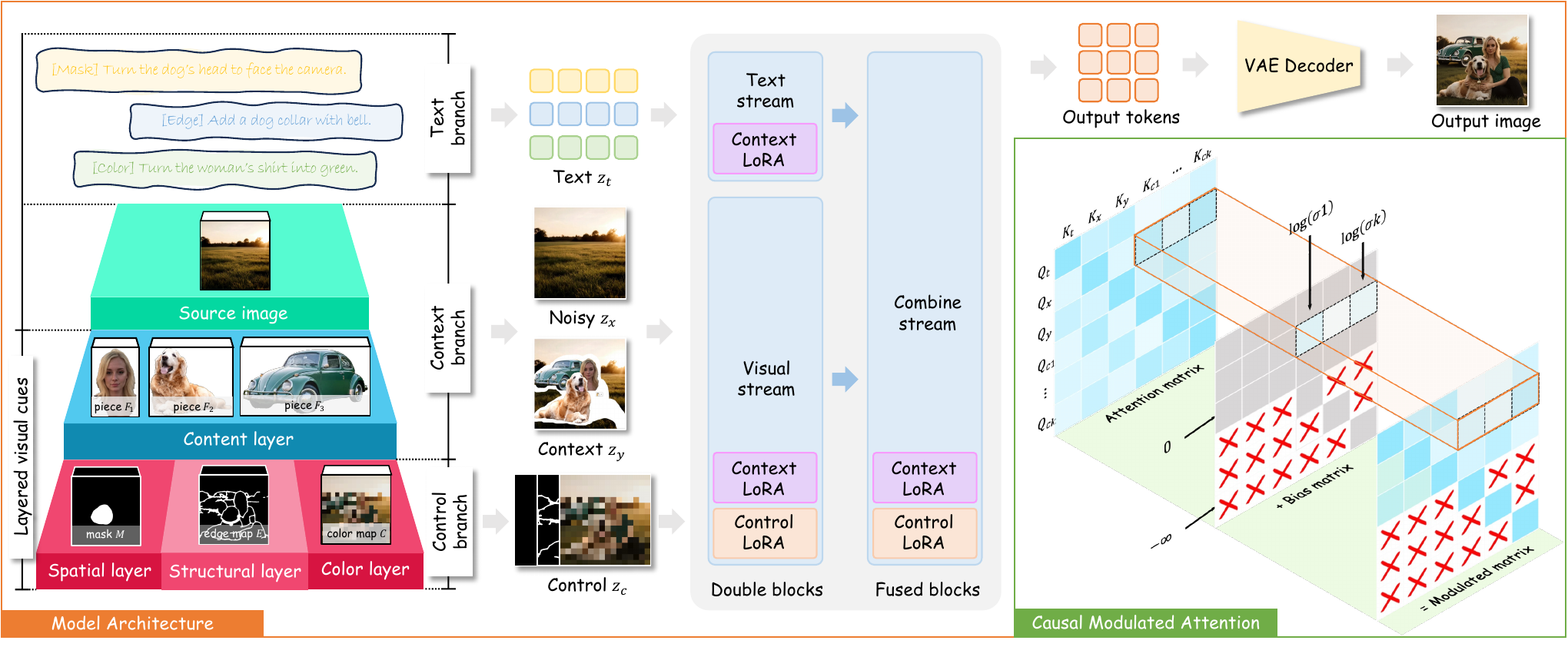}
    \caption{\textbf{Overview of MagicQuill V2 Model Architecture.} Our model processes layered visual cues and text instructions through distinct branches. The layered visual cues (left) are divided into a content layer ($L_{fg}$) and multiple control layers ($L_{control}$: spatial, structural, color), which are encoded into latents ($Z_y$, $Z_c$). These are processed alongside text ($Z_t$) and noisy image ($Z_x$) latents in a unified control module (middle) adapted with dedicated control modules. A Causal Modulated Attention mechanism applies a bias matrix to the attention logits to precisely manage the influence and isolation of each control cue.}
    \label{fig:overview}
\end{figure*}

Our pipeline begins with the synthesis of a base dataset of 5,000 images depicting diverse object-scene interactions. We employ Qwen3-8B \cite{yang2025qwen3} to generate descriptive captions ($c$) and use Flux.1 Krea \cite{fluxkrea2025} to produce high-resolution, photorealistic source images. For each source image, we utilize Grounding SAM \cite{ren2024groundedsam} to extract the primary object's mask, yielding our initial set of foreground pieces and their corresponding source images.

A key challenge is that extracted foregrounds are frequently incomplete due to occlusion (e.g., a hand covering part of an apple). Training on such data would lead to simple copy-paste behavior rather than contextual understanding. To mitigate this, we trained a dedicated object completion model, a LoRA \cite{hu2022lora} fine-tuned on FLUX Kontext \cite{batifol2025kontext}. This model was trained on a dataset of 3,000 complete objects by learning to restore them from randomly applied brushstroke masks. By applying our extracted foregrounds, this LoRA produces a complete and whole version of the foreground object from the previously occluded item.

To ensure robustness against the diverse quality of user-provided inputs, we create what we term the augmented foreground object through a series of augmentations. First, to address photometric inconsistencies, we perform photometric augmentation by applying random lightmaps with ICLight \cite{zhang2025iclight}, forcing the model to learn lighting harmonization. Second, to simulate inputs of varying detail, resolution augmentation involves randomly downsampling and resizing the object. Third, to account for geometric mismatches, perspective augmentation applies a random perspective transformation, introducing plausible distortions.

In the final stage, we assemble the training triplets. The target ($x$) is the unmodified source image, and ($c$) is its descriptive caption. The input image ($y$) is constructed by first compositing the augmented foreground object back into its original position, and then applying random mask augmentations to the surrounding background region, forcing the model to learn contextual harmonization.

To integrate this new capability, we fine-tune the FLUX Kontext \cite{batifol2025kontext} backbone using a Low-Rank Adaptation (LoRA) \cite{hu2022lora} adapter over the attention layer. The model is trained on the triplets ($y, c, x$) generated by our pipeline, optimizing the rectified-flow objective shown below:
{
\begin{equation}
\mathcal{L}_{\theta} = \mathbb{E}_{t \sim p(t), x, y, c} [ \left\| v_{\theta}(z_t, t, y, c) - (\epsilon - x) \right\|_2^2 ],
\end{equation}
}
where $\epsilon \sim \mathcal{N}(0,1)$ and $z_t = (1-t)x + t\epsilon$.

\subsection{Unified Control Module for Control Layers}
\label{subsec:control_layer}
\textbf{Control module.} To seamlessly integrate the diverse cues comprising $L_{control}$, i.e., spatial layer (mask $M$), structural layer (edge map $E$), and color layer (color map $C$), we introduce a minimal and unified control module. 

For computational efficiency, we resize all visual control cues from their original size $(H, W)$ to a fixed low resolution $(h, w)$. To maintain correct spatial alignment, the positional encoding for a resized patch at grid $(i, j)$ is mapped to its original high-resolution coordinates $(P_i, P_j)$ via
$P_i = i \cdot \frac{H}{h}$, $\quad P_j = j \cdot \frac{W}{w}.$

Consider the input representations for different branches: text embedding ($Z_t$), noisy image latent ($Z_x$), and context image latent ($Z_y$). In the Multi-Modal Diffusion Transformer (MMDiT) \cite{flux2023, batifol2025kontext}, the Query-Key-Value (QKV) transformation for image-related branches is defined as:
\begin{equation}
Q_i, K_i, V_i = W_QZ_i, W_KZ_i, W_VZ_i, \quad i\in \{x, y\}
\end{equation}
where $W_Q, W_K, W_V$ are shared projection matrices. 

To incorporate the visual cue latents $Z_c$, we introduce a dedicated conditional branch adapted using Low-Rank Adaptation (LoRA) \cite{hu2022lora}. The updated QKV features for the Condition Branch are computed by adding the LoRA update directly to the standard projection:
\begin{align}
Q_c &= W_QZ_c + B_Q A_Q Z_c, \\
K_c &= W_KZ_c + B_K A_K Z_c, \\
V_c &= W_VZ_c + B_V A_V Z_c,
\end{align}
where $A$ and $B$ are the low-rank decomposition matrices with rank $r \ll d$. The final Query, Key, and Value for Multi-Modal Attention (MMA) can be expressed as:
\begin{align}
\mathbf{Q} &= [Q_t; Q_x; Q_y; Q_c], \\
\mathbf{K} &= [K_t; K_x; K_y; K_c], \\
\mathbf{V} &= [V_t; V_x; V_y; V_c].
\end{align}

To modulate the influence of visual cues on the denoising process, we directly manipulate the attention scores by adding a custom bias matrix, $\mathbf{B}$, to the attention logits:
\begin{equation}
\text{Attention}(\mathbf{Q}, \mathbf{K}, \mathbf{V}) = \text{Softmax}\left(\frac{\mathbf{Q}\mathbf{K}^T}{\sqrt{d_k}} + \mathbf{B}\right) \mathbf{V}.
\label{eq:attention_with_mask}
\end{equation}

This bias matrix $\mathbf{B}$ is designed to selectively manage information flow. Let $\mathcal{I}_t, \mathcal{I}_x, \mathcal{I}_y, \mathcal{I}_{c_k}$ be the token index sets for the text $Z_t$, noisy image $Z_x$, context image $Z_y$, and the $k$-th visual cue $Z_{c_k}$. The bias entry $B_{ij}$ is defined as:
\begin{equation}
B_{ij} =
\begin{cases}
\log(\sigma_k) & \text{if } i \in \mathcal{I}_x \text{ and } j \in \mathcal{I}_{c_k}, \\
-\infty & \text{if } i \in \mathcal{I}_{c_k}, j \notin \mathcal{I}_{c_k}, \\
0 & \text{otherwise.}
\end{cases}
\label{eq:mask_definition}
\end{equation}

The $\log(\sigma_k)$ term acts as a user-adjustable guidance scale, where $\sigma_k \ge 0$ is a scalar parameter modulating the attention from the noisy image latents $Z_x$ to the $k$-th cue $Z_{\sigma_k}$. When $\sigma_k=1$, the bias is $0$, resulting in the standard, unbiased attention mechanism. As $\sigma_k$ increases, the positive bias $\log(\sigma_k)$ strengthens the cue's influence, forcing stricter adherence. Conversely, setting $\sigma_k=0$ makes the bias $-\infty$, effectively disabling the cue. Simultaneously, the second case in Eq. \ref{eq:mask_definition} sets the bias to $-\infty$ for any attention between different control signals to prevent interference.

\noindent\textbf{Training.} We train three distinct control branches for the spatial ($M$), structural ($E$), and color ($C$) layers, which are designed to be used independently or cooperatively. For the structural and color layer ($E,C$), we train the model to generate images from noise conditioned on these cues, rather than on editing pairs. This is achieved by optimizing the rectified-flow objective in Eq. (2), but with $y=\emptyset$ i.e., without the context image $y$. The model thus learns to approximate $p(x | c)$, where the condition $c$ includes the natural language prompt and the specific control map ($E$ or $C$). We found that this capability, learned during conditional generation, robustly generalizes to the conditional editing task at inference time, allowing the model to apply structural or color guidance even when a context image $y$ is provided.

In contrast, the spatial layer (mask $M$) is trained explicitly for local editing, as its purpose is to constrain changes to a specific region. This requires a dataset of (source, target, prompt, mask) tuples, which we generate via self-distillation. First, we use a VLM (Qwen2.5-VL-72B \cite{bai2025qwen2}) to generate several plausible local editing prompts for a given source image. Our base FLUX Kontext model then executes these edits. To derive the mask $M$, we compute the pixel-wise difference between the source and edited images, apply a threshold, and calculate the convex hull of the changed regions. We then filter this dataset to remove samples with masks that are either too large (global edits) or too small (no significant change). To further enhance the model's capability for the common task of object removal, we follow \citet{jiang2025smarteraser} to construct an additional dataset. This involves randomly extracting foreground objects and pasting them back onto arbitrary locations in the same image, strengthening the model's ability to seamlessly remove content.

\begin{figure}[t]
    \centering
    \includegraphics[width=\columnwidth]{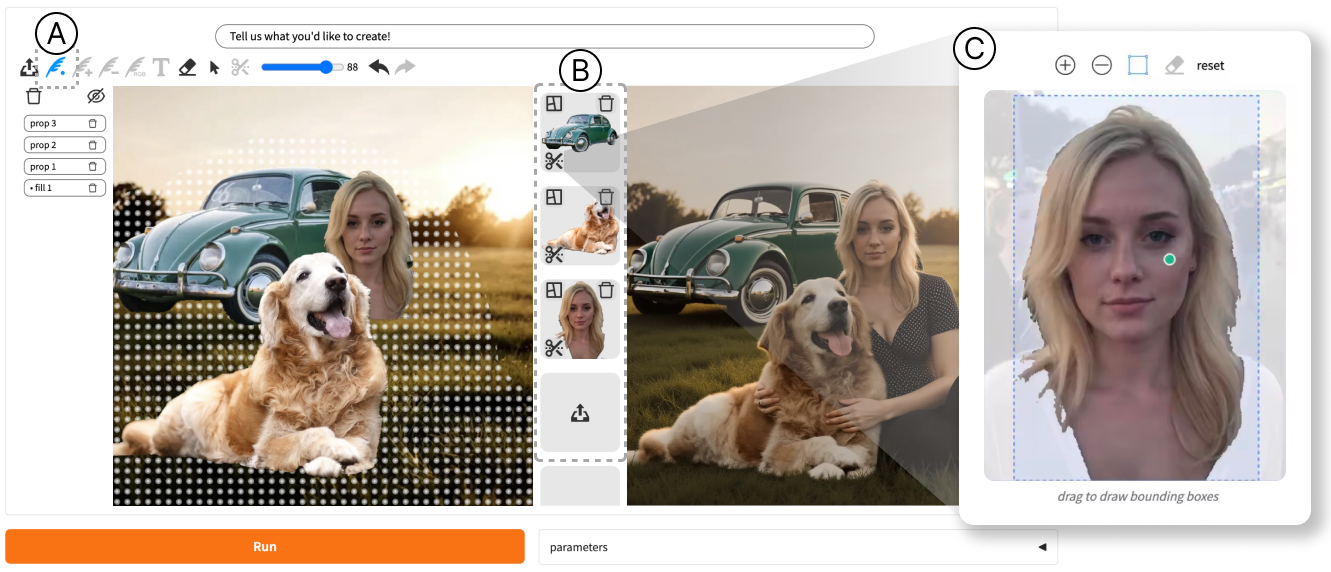}
    \vspace{-15pt}
    \caption{The interactive system interface of MagicQuill V2. (A) The \textit{Fill Brush} in the toolbar allows users to define the spatial layer (mask $M$) by painting on the canvas. (B) The \textit{Visual Cue Manager} holds content layer cues for drag-and-drop composition. (C) The \textit{Image Segmentation Panel}, triggered from the manager, enables precise cue extraction using SAM-based interactions.}
    \label{fig:UI}
    \vspace{-10pt}
\end{figure}

\begin{figure*}[t]
    \centering
    \vspace{-0.2cm}
    \includegraphics[width=\linewidth]{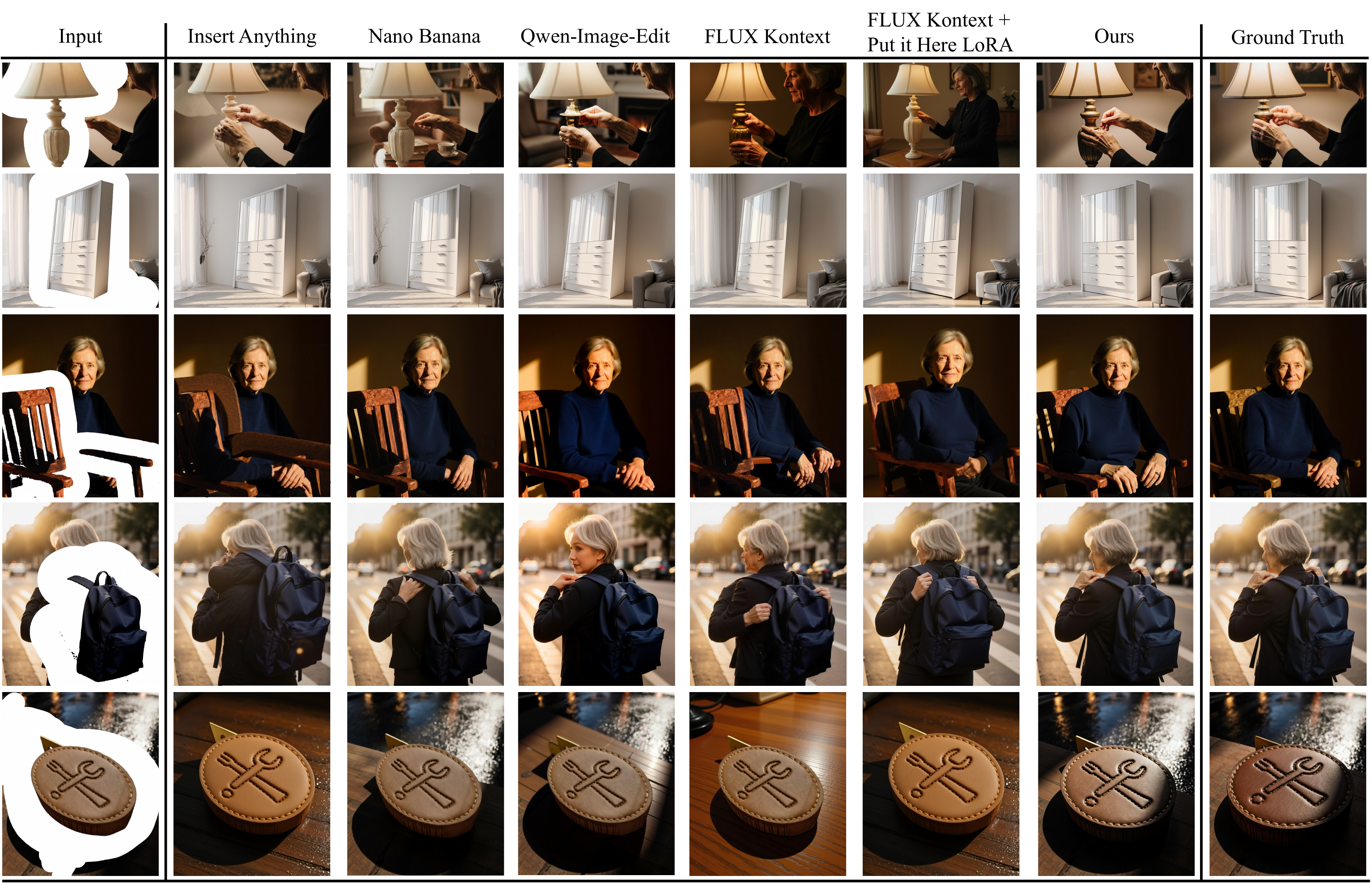}
    \vspace{-0.5cm}
    \caption{Qualitative comparison of our content layer integration against state-of-the-art methods. Our model (MagicQuill V2) successfully integrates foreground objects by handling complex semantic interactions (rows 3, 4), harmonizing lighting (row 1), and correcting perspective distortions (rows 2, 5). These examples highlight our method's ability to produce results that are significantly more realistic and faithful than Insert Anything \cite{song2025insert_anything}, Nano Banana \cite{nano_banana}, Qwen-Image \cite{bai2025qwen2}, and the specialized Kontext LoRA ``Put it Here'' \cite{bai2025qwen2}.}
    \label{fig:comparison}
    \vspace{-0.3cm}
\end{figure*}

\subsection{Interactive System Interface}
\label{subsec:UI}
We designed an interactive system, shown in~\cref{fig:UI}, to unify our layered composition framework. This interface extends the ``Idea Collector'' from MagicQuill V1  \cite{liu2025magicquill}. The main Canvas is supplemented by a \textit{Toolbar} containing a new \textit{Fill Brush} (A). This tool is the primary interaction for defining the spatial layer (mask $M$), allowing users to simply paint where the edit should occur. 
The Toolbar also retains the similar brush-based tools from V1, enabling users to draw sketches for the structural layer (edge map $E$) and apply strokes for the color layer (color map $C$).
A \textit{Visual Cue Manager} (B), located right to the canvas, holds all content layer ($L_{fg}$) visual cues (foreground pieces $F_i$). Users can drag these cues onto the canvas to define what to generate.

To help users extract these visual cues from source images, each cue in the \textit{Visual Cue Manager} features a ``segment'' button, which opens an \textit{Image Segmentation Panel} (C). This panel utilizes SAM \cite{kirillov2023sam} to precisely segment an object in real-time, using positive/negative dots or a bounding box. The refined cue can then be saved back to the manager. This workflow provides an intuitive method for managing and composing precise visual cues.

\section{Experiment}
We conduct a comprehensive set of experiments to validate the effectiveness of our proposed layered composition framework. Our evaluation is structured to first analyze the fidelity and contextual awareness of the content layer (\cref{subsec:analysis_content}), followed by a detailed assessment of the precision and versatility of the control layers (\cref{subsec:analysis_control} and \cref{subsec:analysis_spatial}). Implementation details and more results will be shown in the supplementary materials.

\subsection{Analysis of Content Layer}
\label{subsec:analysis_content}
The goal of the content layer ($L_{fg}$) is to perform context-aware edits around the user-provided foreground cues, seamlessly integrating the foreground pieces into the edited image, rather than a simple ``copy-paste'' operation. Following the principles of our data construction pipeline detailed in \cref{subsec:content_layer}, we manually curate a test set of $200$ samples. This set is designed to be challenging and is divided into two categories: 100 interaction-based samples and 100 placement-based samples. We evaluate our method against a suite of state-of-the-art models, including InsertAnything \cite{song2025insert_anything}, Nano Banana \cite{nano_banana}, Qwen-Image \cite{wu2025qwen-image}, the base FLUX Kontext model \cite{batifol2025kontext}, and a popular, community-trained Kontext LoRA ``Put it Here'' \cite{putitherev4} specifically designed for high-fidelity foreground insertion. 

\begin{figure*}[t]
    \centering
    \vspace{-0.2cm}
    \includegraphics[width=\linewidth]{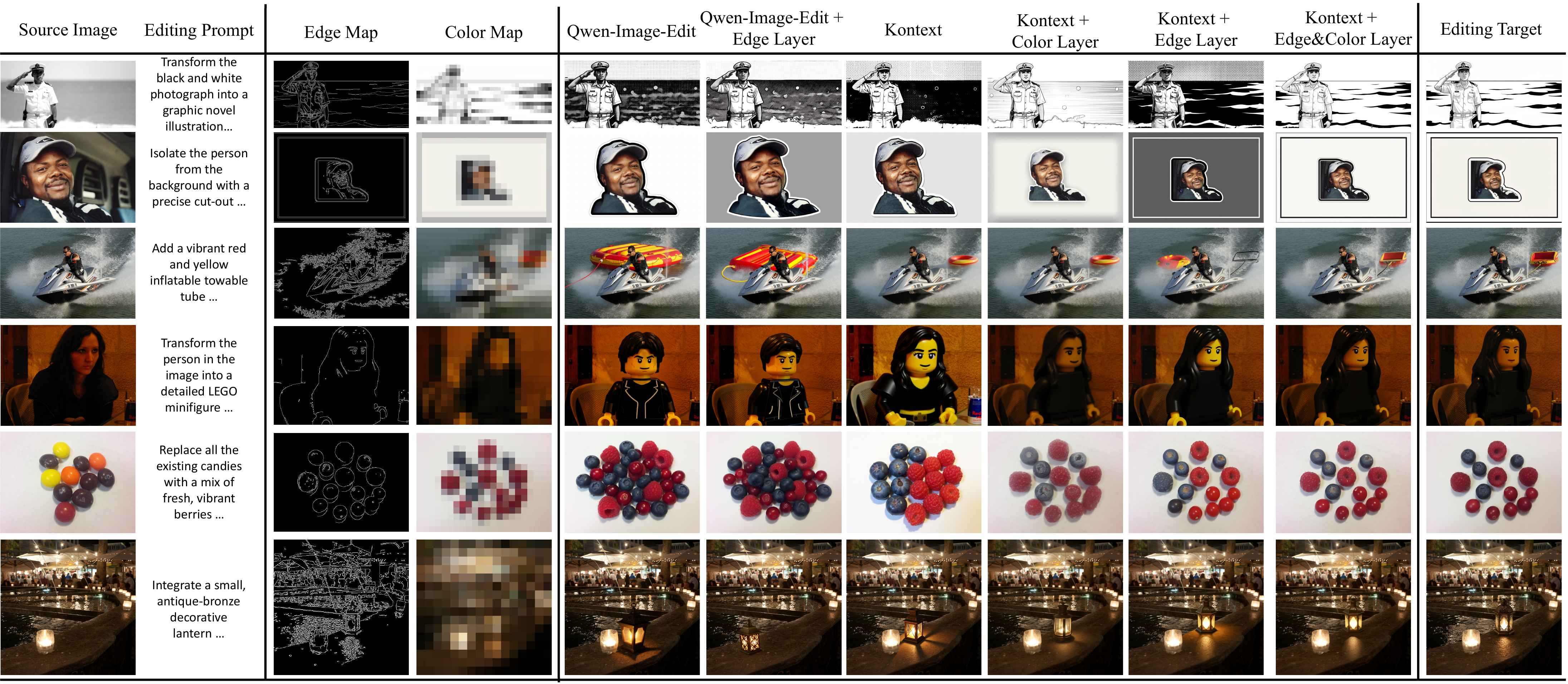}
    \vspace{-0.7cm}
    \caption{Qualitative comparison for structural (edge) and color layer control. Our full model, by composing both cues, achieves high-precision edits, significantly outperforming baselines.}
    \label{fig:comparison_control}
    \vspace{-0.3cm}
\end{figure*}

\noindent\textbf{Qualitative and quantitative comparison.}
\cref{fig:comparison} provides a comprehensive visual comparison. Our model excels at complex interactions, realistically generating the hand around the backpack (row 4) and the person sitting on the chair (row 3). Furthermore, our relight augmentation leads to superior photometric harmonization, as seen in the lamp and hand's illumination (row 1) and logo's correct side-lighting (row 4). The successful geometric correction of the cabinet (row 2) and logo (row 5) validates our use of perspective augmentation. 

In contrast, other baselines struggle with these scenarios, altering non-edited regions (row 1) or failing complex interactions (rows 3-4), lighting (row 5), and geometry (row 2). Our method's consistent success validates that our data pipeline focuses on foreground completion, and augmentations are essential for robust, high-fidelity synthesis. This is confirmed quantitatively in \cref{tab:content_quant}, where our method significantly outperforms all baselines on almost all metrics.


\begin{table}[t] 
\centering 
\small
\vspace{-6pt}
\caption{
    Quantitative comparison for the content layer composition. Our model achieves the best in almost all metrics.
}
\label{tab:content_quant}
\vspace{-10pt}
\SetTblrInner{rowsep=0.0pt}
\SetTblrInner{colsep=1.0pt} 
\begin{tblr}{
    cells = {halign=c, valign=m},   
    column{1} = {halign=l},          
    column{2,3} = {colsep=4pt},
    cell{1}{1} = {font=\bfseries},   
    hline{1,Z} = {1-7}{1.0pt},     
    hline{2} = {1-7}{0.5pt},      
    vline{2} = {1-Z}{0.5pt},      
}
Model & $L_1 \downarrow$ & $L_2 \downarrow$ & CLIP-I $\uparrow$ & DINO $\uparrow$ & CLIP-T $\uparrow$ & LPIPS $\downarrow$ \\
Insert Anything & 0.105 & 0.039 & 0.910 & 0.825 & 0.327 & 0.354 \\
Nano Banana & 0.105 & 0.038 & 0.934 & 0.891 & 0.335 & 0.321 \\
Qwen-Image & 0.114 & 0.042 & 0.929 & 0.881 & 0.334 & 0.357 \\
FLUX Kontext & 0.117 & 0.045 & 0.930 & 0.872 & \textbf{0.337} & 0.359 \\
Put it Here & 0.136 & 0.054 & 0.925 & 0.854 & 0.335 & 0.438 \\
\textbf{Ours} & \textbf{0.061} & \textbf{0.019} & \textbf{0.962} & \textbf{0.930} & 0.335 & \textbf{0.202} \\
\end{tblr}
\vspace{-10pt} 
\end{table}

\begin{figure}[t]
    \centering
    \includegraphics[width=\columnwidth]{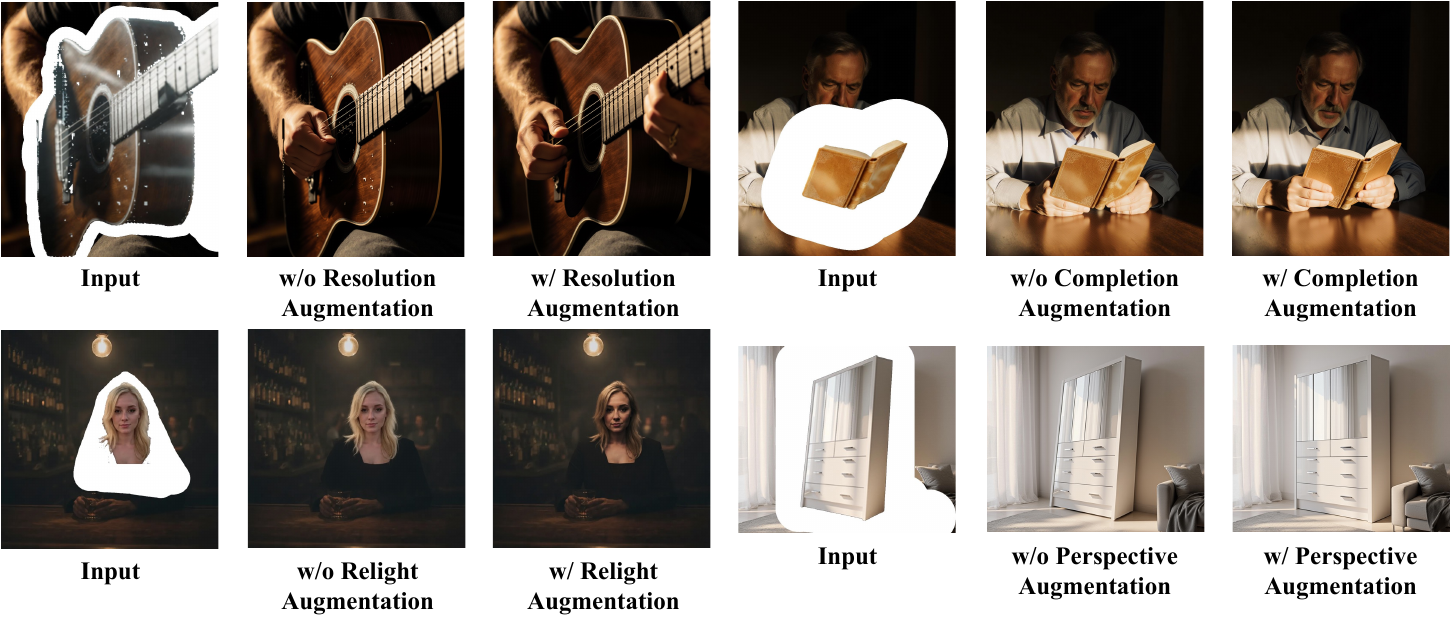}
    \vspace{-20pt}
    \caption{Qualitative ablation of our data construction pipeline (\cref{subsec:content_layer}). Removing Perspective, Relight, Resolution, or Object Completion augmentations leads to synthesis failures.}
    \label{fig:data_ablation}
    \vspace{-15pt}
\end{figure}

\noindent\textbf{Ablation in Data Construction.} We conduct an ablation study to validate the individual components of our data construction pipeline, detailed in \cref{subsec:content_layer}. We train separate models, each lacking one of the key augmentations, and present the qualitative results in \cref{fig:data_ablation}.

Without training on perspective-distorted inputs, the model fails to correct the geometry. As seen, it places the cabinet with its original slant, resulting in a physically implausible scene (row 1). Without relight augmentation, the model defaults to a ``pasted-on'' look. The subject's lighting is flat and does not match the bar's warm, ambient lighting (row 2). When robustness to varied input quality is not learned, the model struggles with low-resolution or low-quality cues, producing a blurry or noisy image. Without our completion LoRA, the model is trained on incomplete foregrounds (e.g., the occluded book). Consequently, it learns a simple ``copy-paste'' behavior and fails to interact with the foreground. 

\subsection{Analysis of Structural and Color Layer}
\label{subsec:analysis_control}
To evaluate the structural layer $L_{structural}$ (edge) and the color layer $L_{color}$ (color), we conduct our evaluation on the Pico-Banana-400K benchmark \cite{qian2025picobanana400klargescaledatasettextguided}, which provides a diverse set of text–image–edit triplets consisting of 35 edit operations across 8 semantic categories. Images from selected benchmark were generated by the state-of-the-art Nano Banana \cite{nano_banana} model and verified by Gemini \cite{gemini2.5} to ensure high fidelity and perfect semantic alignment. 

To test MagicQuill V2's precision and its ability to follow control cues, we sampled 1,000 cases from this benchmark. For each triplet, we simulate a user with a target in mind by extracting the edge map and color map directly from the target image. We then provide these cues to the models. We compare against the latest open-source model Qwen-Image-Edit \cite{wu2025qwen-image}, which includes native edge control module. We also show ablations of our own model with edge and color control layers.

\noindent\textbf{Qualitative and quantitative comparison.} The qualitative results in \cref{fig:comparison_control} demonstrate the remarkable precision of our model. Qwen-Image-Edit \cite{wu2025qwen-image} and FLUX Kontext \cite{batifol2025kontext} interpret the instruction to generate plausible edit. However, text prompts alone are often insufficient to capture complex spatial and color intentions. The Edge module of Qwen-Image-Edit attempts to address this, but its control remains `soft' and interpretive, failing to align perfectly. 

In contrast, our layered approach highlights the complementary value of different control modalities. Our model with the edge layer alone perfects geometry but distorts color, while the color layer alone aligns color but misses structural details. This proves both cues are essential. When composed together, their strengths are combined to perfectly reproduce the user's intent, consistently matching the target edit. This validates that our layered design provides the crucial bridge to high-fidelity, explicit control. These qualitative findings are directly supported by our quantitative evaluation in \cref{tab:control_quant}, where our combined method performs uniformly better across all metrics.

\begin{table}[t] 
\centering 
\small
\vspace{-6pt}
\caption{
    Quantitative comparison for the control layers.
}
\label{tab:control_quant} 
\vspace{-10pt}
\SetTblrInner{rowsep=0.0pt}
\SetTblrInner{colsep=1.0pt} 
\begin{tblr}{
    cells = {halign=c, valign=m},   
    column{1} = {halign=l},          
    column{2,3} = {colsep=4pt}, 
    cell{1}{1} = {font=\bfseries},   
    hline{1,Z} = {1-6}{1.0pt},     
    hline{2} = {1-6}{0.5pt},      
    vline{2} = {1-Z}{0.5pt},      
}
Model & $L_1 \downarrow$ & $L_2 \downarrow$ & CLIP-I $\uparrow$ & DINO $\uparrow$ & LPIPS $\downarrow$ \\
Qwen Image & 0.132 & 0.043 & 0.923 & 0.871 & 0.395 \\
Qwen Image (Edge) & 0.131 & 0.042 & 0.924 & 0.875 & 0.387 \\
FLUX Kontext & 0.152 & 0.054 & 0.908 & 0.853 & 0.434 \\
Ours (Edge) & 0.107 & 0.030 & 0.938 & 0.909 & 0.317 \\
Ours (Color) & 0.080 & 0.020 & 0.943 & 0.915 & 0.327 \\
\textbf{Ours (Edge$+$Color)} & \textbf{0.080} & \textbf{0.018} & \textbf{0.949} & \textbf{0.930} & \textbf{0.283}
\end{tblr}
\vspace{-12pt} 
\end{table}

\begin{figure}[t]
    \centering
    \includegraphics[width=\columnwidth]{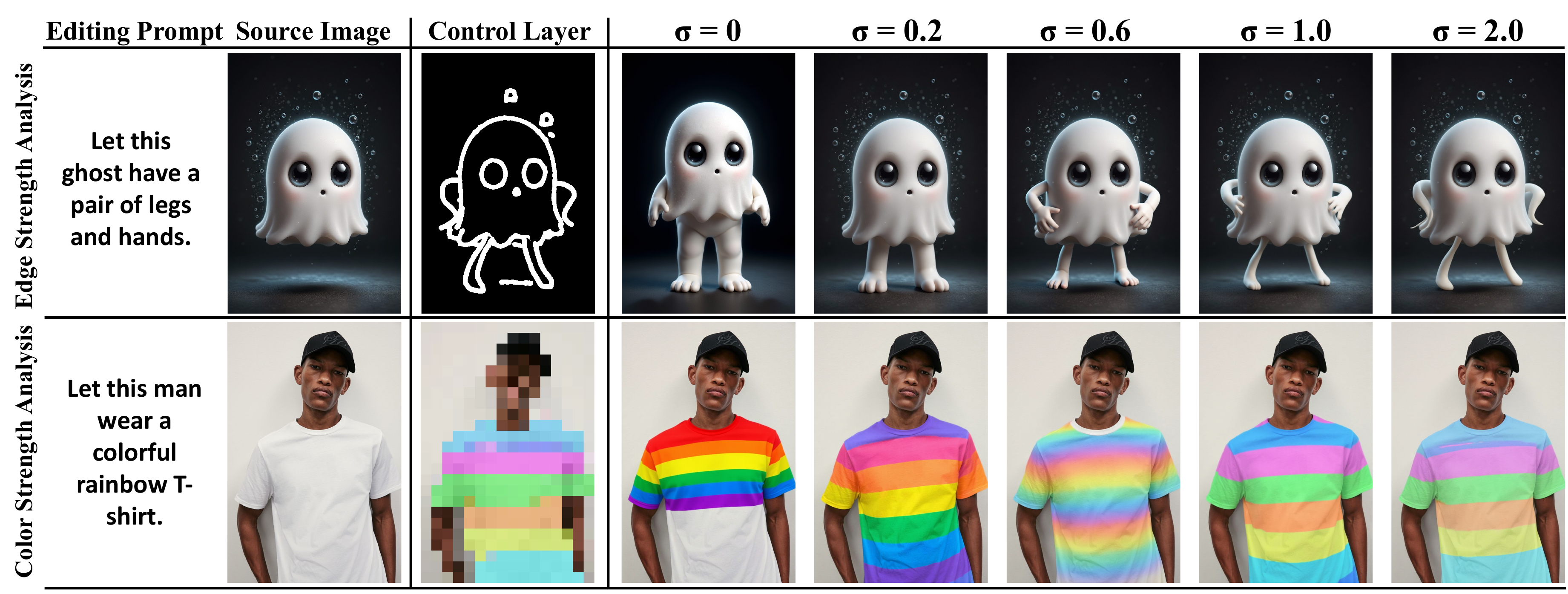}
    \vspace{-20pt}
    \caption{Qualitative analysis of the control strength parameter $\sigma$. As $\sigma$ increases, the model's adherence to the user-provided edge layer (row 1) and color layer (row 2) progressively tightens.}
    \label{fig:control_strength}
    \vspace{-15pt}
\end{figure}

\noindent\textbf{Analysis of control strength.} The control strength parameter $\sigma$ (\cref{eq:mask_definition}) offers flexible control, which is crucial as user-provided cues, such as hand-drawn sketches, can be rough. Users can tune $\sigma$ to balance their trust in the cue against the model's generative priors. We analyze this mechanism in \cref{fig:control_strength}, using manually drawn cues to simulate this scenario. At $\sigma = 0$, the cue is ignored, reverting to the base FLUX Kontext behavior. As $\sigma$ increases, adherence to the user-drawn edge layer (row 1) and color layer (row 2) progressively tightens. While our default $\sigma = 1.0$ provides a balanced and faithful result, higher values enforce stricter adherence, which can also amplify imperfections in the cue, potentially leading to artifacts.

\subsection{Analysis of Spatial Layer} 
\label{subsec:analysis_spatial}
In this section, we evaluate the spatial layer ($L_{spatial}$), which is designed for precise regional edit. We first compare its ability to perform local edits against inpainting-based models. We then evaluate its specialized object removal capability, for which it was explicitly fine-tuned.

\begin{figure}[t]
    \centering
    \includegraphics[width=\linewidth]{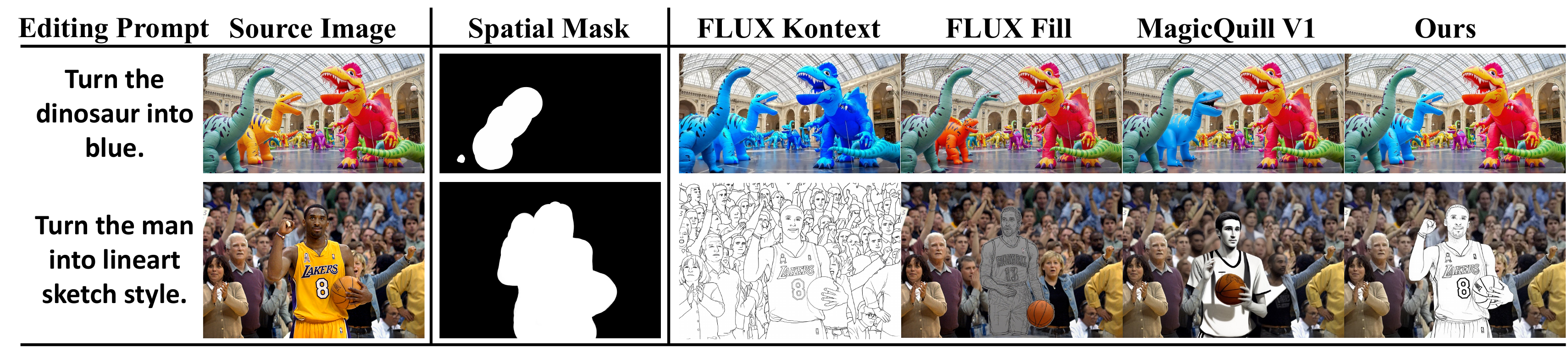}
    \vspace{-20pt}
    \caption{Qualitative comparison for regional editing.}
    \vspace{-12pt}
    \label{fig:comparison_inpainting}
\end{figure}

\noindent\textbf{Comparison with inpainting models.} Our spatial layer is designed to apply edits to masked content, not just regenerate. As shown in \cref{fig:comparison_inpainting}, FLUX Kontext fails with global edits, while inpainting models like FLUX Fill \cite{flux2023} and MagicQuill V1 \cite{liu2025magicquill} ignore the masked region's content, failing content-aware edits (e.g., Row 2). Our model, however, attends to the content within the mask, successfully applying color adjustment and style transfer while preserving the identity, demonstrating content-aware local editing.

\begin{figure}[t]
    \centering
    \includegraphics[width=\linewidth]{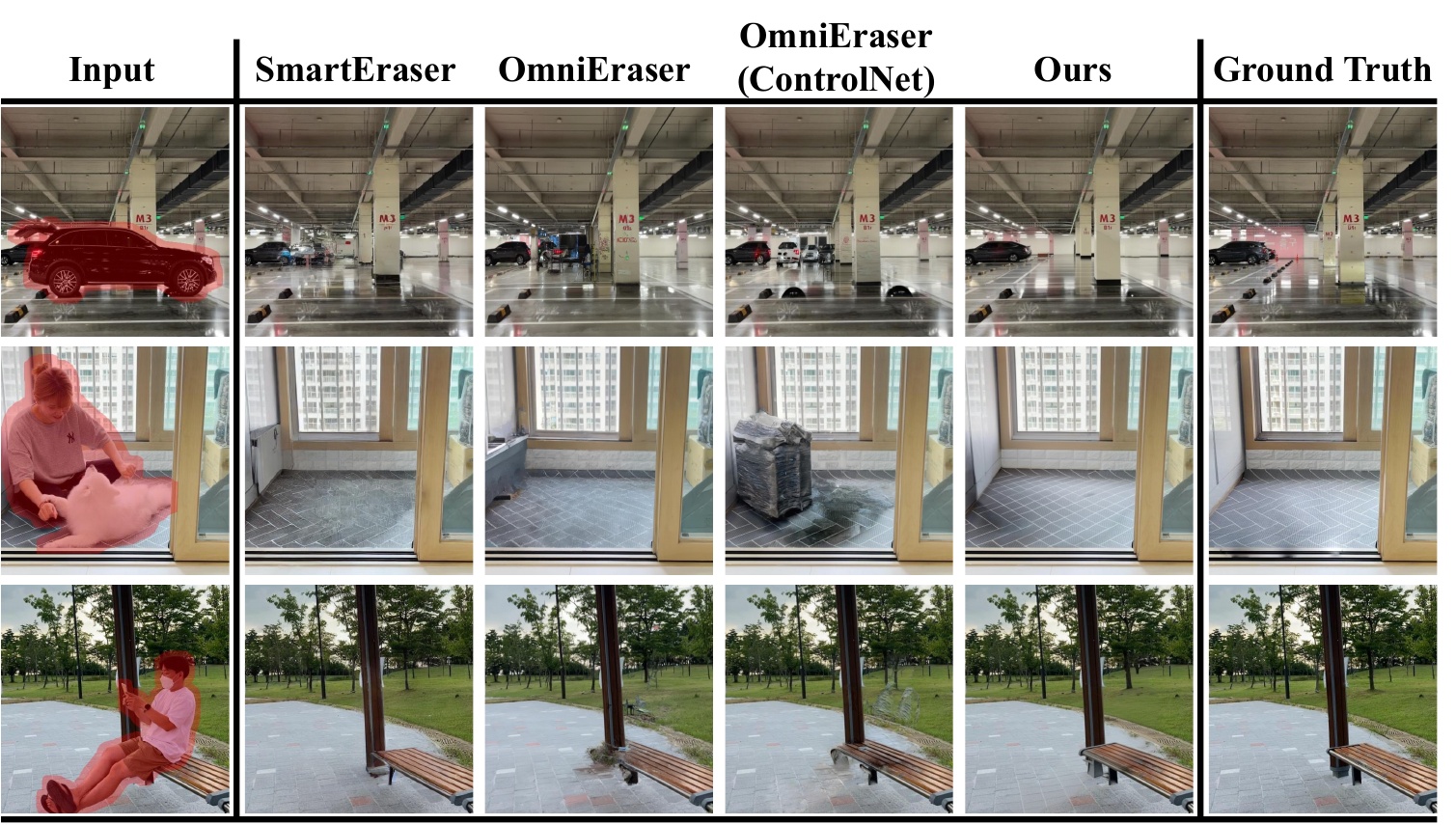}
    \vspace{-20pt}
    \caption{Qualitative comparison for object removal task.}
    \vspace{-15pt}
    \label{fig:object_removal}
\end{figure}

\noindent\textbf{Comparison in object removal.} We compare our model with state-of-the-art object removal methods SmartEraser \cite{jiang2025smarteraser} and OmniEraser \cite{wei2025omnieraser} over 5,000 samples of RORD \cite{sagong2022rord} benchmark. The experimental results are presented in~\cref{tab:object_removal} and~\cref{fig:object_removal}. MagicQuil V2 demonstrates superior performance over existing models in erasing target objects.

\begin{table}[t] 
\centering 
\small
\vspace{-6pt}
\caption{
    Quantitative comparison for object removal.
}
\label{tab:object_removal}
\vspace{-10pt}
\SetTblrInner{rowsep=0.0pt}
\SetTblrInner{colsep=1.0pt} 
\begin{tblr}{
    cells = {halign=c, valign=m},   
    column{1} = {halign=l},          
    column{2,3} = {colsep=4pt},
    cell{1}{1} = {font=\bfseries},   
    hline{1,Z} = {1-7}{1.0pt},     
    hline{2} = {1-7}{0.5pt},      
    vline{2} = {1-Z}{0.5pt},      
}
Model & $L_1 \downarrow$ & $L_2 \downarrow$ & LPIPS $\downarrow$ & SSIM $\uparrow$ & PSNR $\uparrow$ & FID $\downarrow$\\
SmartEraser & 0.069  & 0.098  &  0.196 &  0.630 &  21.14  & 17.03 \\
OmniEraser (Base) & 0.058  & 0.084  &  0.243 & 0.660  &  22.16  & 19.76 \\
OmniEraser (CN) & 0.048  &  0.084  &  0.182 &  0.817 &  22.96  & 25.92 \\
\textbf{Ours} & \textbf{0.042} & \textbf{0.071} & \textbf{0.154} &  \textbf{0.840} & \textbf{24.45} & \textbf{16.42} \\
\end{tblr}
\vspace{-10pt} 
\end{table}

\section{Conclusion}
We introduced MagicQuill V2, a novel system that brings the granular control of layered composition to generative image editing. Monolithic text prompts are insufficient for capturing complex user intent. Our method deconstructs this intent into distinct content, spatial, structural, and color layers. We proposed a specialized data pipeline for context-aware content layer composition and a unified control module to precisely manage geometry, color, and regional edits. Extensive experiments validate our approach, demonstrating state-of-the-art performance in content composition, alignment with respect to the control layer, and high-fidelity regional edit. MagicQuill V2 effectively bridges the intention gap, granting users intuitive, powerful, and precise control over the generative process.

{
    \small
    \bibliographystyle{ieeenat_fullname}
    \bibliography{main}
}

\clearpage
\setcounter{page}{1}
\maketitlesupplementary

\section{Implementation Details}
\label{sec:Implementation}
\subsection{Content Layer} 
\noindent\textbf{Training and Inference Details.}  We integrate the content layer capability by fine-tuning the FLUX Kontext backbone with a LoRA adapter of rank $r=32$. The model was trained for 9,000 steps on 8 H20 (140GB) GPUs, using the AdamW optimizer with a constant learning rate of $1 \times 10^{-4}$. For inference, the model performs 20 denoising steps, requiring approximately 30 seconds and 30GB of VRAM on a single H20 GPU.

\begin{figure}[h]
    \centering
    \includegraphics[width=0.8\linewidth]{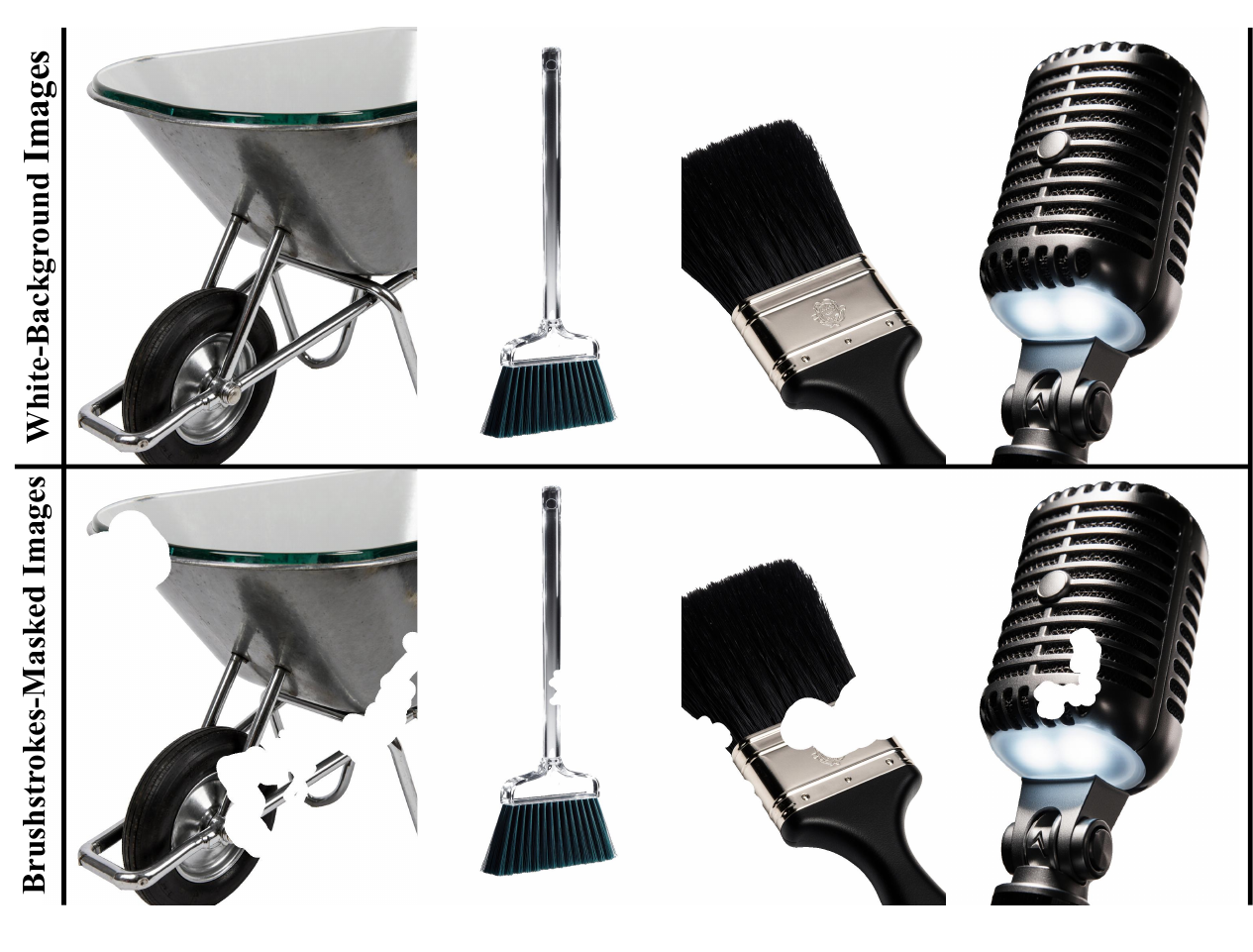}
    \caption{Training data generation for the completion LoRA.}
    \vspace{-0.4cm}
    \label{fig:appen_complete_lora}
\end{figure}

\noindent\textbf{Completion Augmentation.} We utilize a dedicated object completion LoRA to restore foreground objects that are occluded in the source images. To train this completion model, we constructed a dataset of 3,000 high-resolution object images with white backgrounds, generated using Flux.1 Krea \cite{fluxkrea2025}.
We simulate occlusions by applying random white brushstroke masks to these images. As demonstrated in \cref{fig:appen_complete_lora}. We train the model by providing pairs of complete, white-background objects (top row) and their counterparts with random brushstroke masks applied (bottom row).

\begin{figure}[h]
    \centering
    \includegraphics[width=0.8\linewidth]{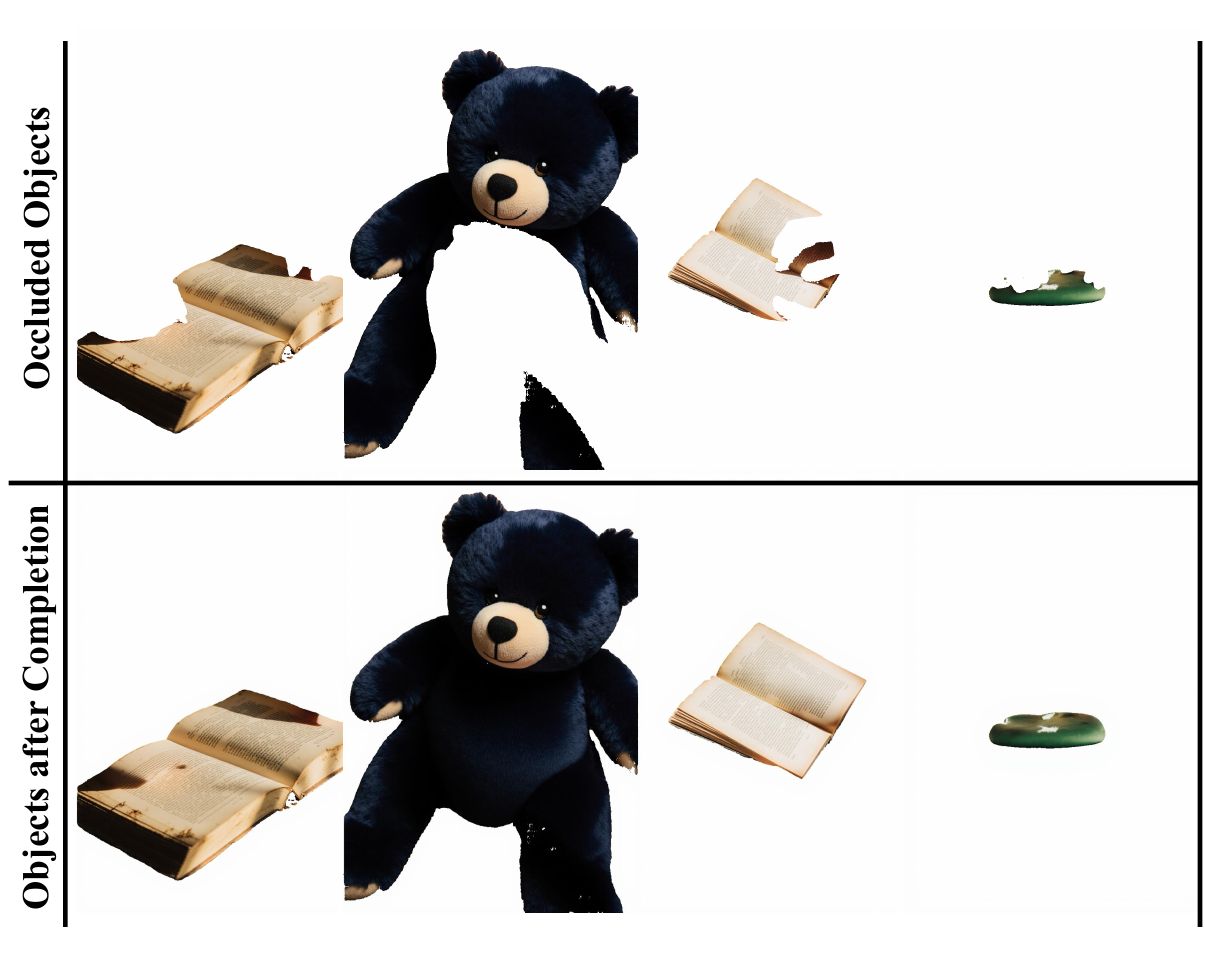}
    \vspace{-0.4cm}
    \caption{Qualitative results of the completion LoRA. }
    \vspace{-0.4cm}
    \label{fig:appen_complete_obj}
\end{figure}

The LoRA is then fine-tuned on FLUX Kontext, with a rank $r=32$, to reconstruct the original complete object from the masked input, guided by its descriptive caption. The training was conducted on 8 H20 (140GB) GPUs. We used the AdamW optimizer with a learning rate of $1 \times 10^{-4}$ with 10 epochs of optimization. Qualitative results of this completion model are shown in \cref{fig:appen_complete_obj}, demonstrating its ability to restore occluded items extracted from scenes.

\begin{figure}[h]
    \centering
    \includegraphics[width=0.8\linewidth]{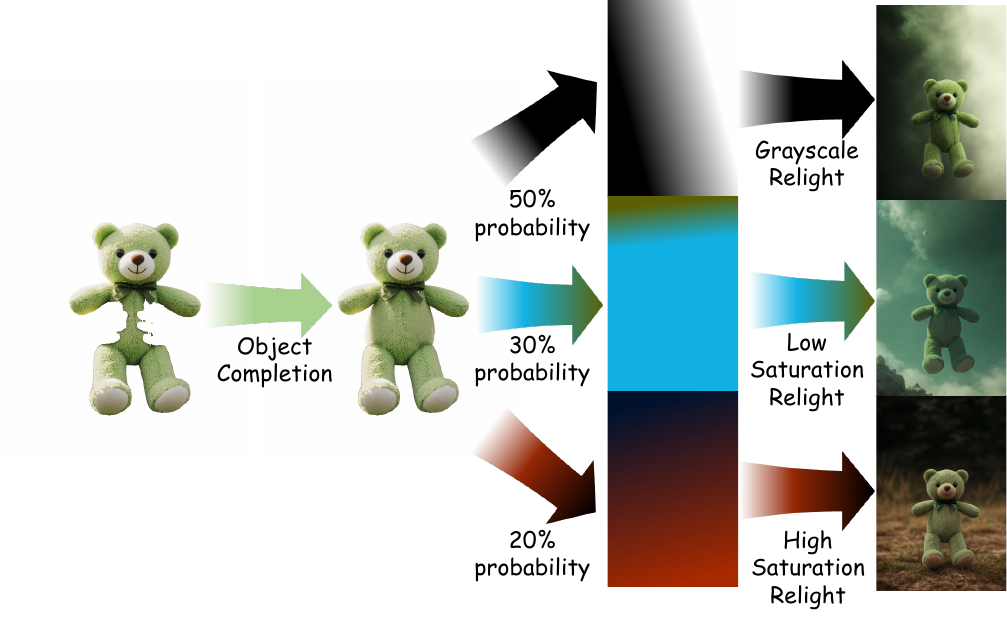}
    \vspace{-0.4cm}
    \caption{Overview of the relight augmentation pipeline. Completed objects are randomly relit using one of three light map categories to improve photometric robustness.}
    \vspace{-0.4cm}
    \label{fig:appen_relight}
\end{figure}

\noindent\textbf{Relight Augmentation.} We perform photometric augmentation using the SD1.5 \cite{rombach2022SD} version of ICLight \cite{zhang2025iclight}. For each foreground object, we first generate a random light map. This map is used as the initial latent in an image-to-image relighting process. The light map's color properties are randomly sampled: 50\% are grayscale, 30\% are low-saturation color, and 20\% are high-saturation color, as illustrated in \cref{fig:appen_relight}. This forces the model to learn robust lighting harmonization rather than just copy-pasting. 

\noindent\textbf{Perspective Augmentation.} To simulate geometric mismatches, we apply a random perspective transformation. Given a foreground object $F$ within its $w \times h$ bounding box, we define its four corner coordinates as the source points $P_{src} = \{ (0, 0), (w, 0), (w, h), (0, h) \}$. We sample a maximum perturbation ratio $\rho \sim U(0.1, 0.3)$ and define the maximum displacement values $\Delta x = w \cdot \rho$ and $\Delta y = h \cdot \rho$. We then generate four destination points $P_{dst} = \{ (x'_i, y'_i) \}_{i=1}^4$ by adding a random displacement $\delta \sim U([-\Delta x, -\Delta y], [\Delta x, \Delta y])$ to each corresponding source point $p_i \in P_{src}$, clipping the result to stay within the original $w \times h$ boundaries.
A $3 \times 3$ perspective transformation matrix $H$ is computed by solving the homography that maps $P_{src}$ to $P_{dst}$. This transformation $H$ is then applied to both the foreground object $F$ and its mask $M_F$ to produce the geometrically augmented $F'$ and $M'_{F}$.

\noindent\textbf{Resolution Augmentation.} To ensure robustness against varying input qualities, we simulate low-resolution inputs. For each foreground object, we sample a random scaling factor $s \sim U(0.15, 0.9)$. The object is downsampled to $s$ of its original size and then upsampled back to its original resolution using bilinear interpolation.

\subsection{Control Layer}
\noindent\textbf{Training and Inference Details.} We implement the structural, color, and spatial layers using the unified control module, each with a rank of $r=128$. All visual control cues (e.g., edge maps, color maps) are resized to a fixed 512x512 resolution before being processed. The control modules were trained for about 10,000 steps on 8 H20 (140GB) GPUs, using the AdamW optimizer with a constant learning rate of $1 \times 10^{-4}$. For inference, the model performs 20 denoising steps. When a control cue is active, this process requires approximately 45 seconds and 40GB of VRAM on a single H20 GPU.
These control modules can be composed and activated simultaneously (e.g., using structural and color cues together for high-fidelity edits) and are also compatible with other LoRAs over FLUX Kontext weights. Following the interactive paradigm of MagicQuill V1 \cite{liu2025magicquill}, our system automatically selects and provides the appropriate control cues based on the user's brushstroke type.

\begin{figure}[h]
    \centering
    \vspace{-0.4cm}
    \includegraphics[width=1.0\linewidth]{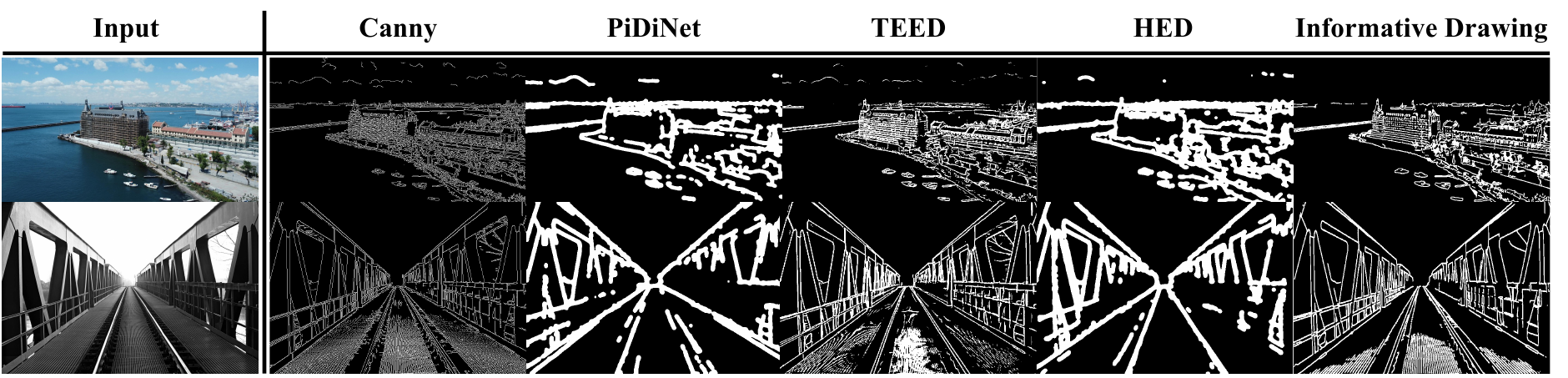}
    \vspace{-0.4cm}
    \caption{Visual results of different edge/lineart extractors. }
    \vspace{-0.4cm}
    \label{fig:appen_edge}
\end{figure}

\noindent\textbf{Structural Layer.} To train the structural layer, we use a large-scale, self-collected dataset, with all images resized to approximately 1000x1000 pixels with detailed captions. For each image, we generate a structural map by randomly selecting one edge and lineart extractor from Canny \cite{ding2001canny}, PidiNet \cite{su2021pidinet}, TEED \cite{Soria_2023teed}, HED \cite{xie2015hed}, and Informative Drawings \cite{chan2022drawings}. As illsutrated in \cref{fig:appen_edge}, this ensures the model is robust to various styles of structural inputs.

The structural layer is trained for conditional generation, optimizing the model to approximate $p(x|c, E)$, where $c$ is the text caption and $E$ is the sampled edge map without the context image. We found that the structure-following capability, learned via conditional generation, robustly generalizes to the conditional editing task at inference time. Following the interactive paradigm of MagicQuill V1 \cite{liu2025magicquill}, we define binary masks $M_{add}$ and $M_{sub}$ corresponding to the user's add and subtract brushstrokes, respectively. The subtract brush removes edges from the map, while the add brush introduces new edges. Let $E$ denote the extracted edge map; the final control condition $E_{cond}$ is formally computed as:
{\small
\begin{equation}
\begin{split}
&\mathbf{E}_{sub} = \mathbf{E} \odot (1 - \mathbf{M}_{sub}), \\
&\mathbf{E}_{cond} = \mathbf{E}_{sub} + \mathbf{M}_{add} \odot (1 - \mathbf{E}_{sub}).
\end{split}
\end{equation}
}
\noindent where $\odot$ denotes the element-wise product. The resulting modified edge map $E_{cond}$ serves as the precise geometric constraint for the generation process.

\noindent\textbf{Color Layer.} The training paradigm for the color layer mirrors that of the structural layer. We utilize the same dataset and train the model for conditional generation without the context image context $y$. During training, the color condition map $\mathbf{C}$ is generated by downsampling to $16\times16$ and resizing back to $512\times512$. This removes high-frequency details, forcing the model to learn the mapping from the palette to the final image.

At inference, we adopt the color brushstroke logics from MagicQuill V1 \cite{liu2025magicquill}. Each user color stroke is parameterized as a tuple $(\mathbf{M}_{color}, \mathbf{c}, \alpha)$, where $\mathbf{M}_{color}$ denotes the binary mask of the painted region, $\mathbf{c}$ specifies the target RGB color, and $\alpha \in [0, 1]$ (default to $0.4$) represents the stroke opacity. Let $\mathbf{C}$ denote the initial color map extracted from the source image (or a blank canvas); the updated control condition $\mathbf{C}_{cond}$ is computed via alpha blending:
{\small
\begin{equation}
\mathbf{C}_{cond} = (1 - \alpha \cdot \mathbf{M}_{color}) \odot \mathbf{C} + \alpha \cdot \mathbf{M}_{color} \cdot \mathbf{c},
\end{equation}
}
\noindent where the specific color $\mathbf{c}$ is applied over the region defined by $\mathbf{M}_{color}$ with intensity $\alpha$. This blending mechanism allows users to achieve both subtle color tinting and opaque color replacement.

\noindent\textbf{Spatial Layer.} Unlike the structural and color layers, which can be trained on single images via conditional generation, the spatial layer (represented by a binary mask $\mathbf{M}$) requires training on local editing pairs $(I_{src}, I_{tgt}, c, \mathbf{M})$. The goal is to constrain the model's generative changes strictly within the user-specified region. To achieve this, we construct a large-scale dataset via a rigorous Self-Distillation Pipeline. We leverage Qwen2.5-VL-72B \cite{bai2025qwen2} to propose a set of plausible local editing instructions $c$ for each $I_{src}$, which are then executed by the base FLUX Kontext \cite{batifol2025kontext} model to produce the edited target image $I_{tgt}$.

\begin{figure}[t]
    \centering
    \vspace{-0.4cm}
    \includegraphics[width=1.0\linewidth]{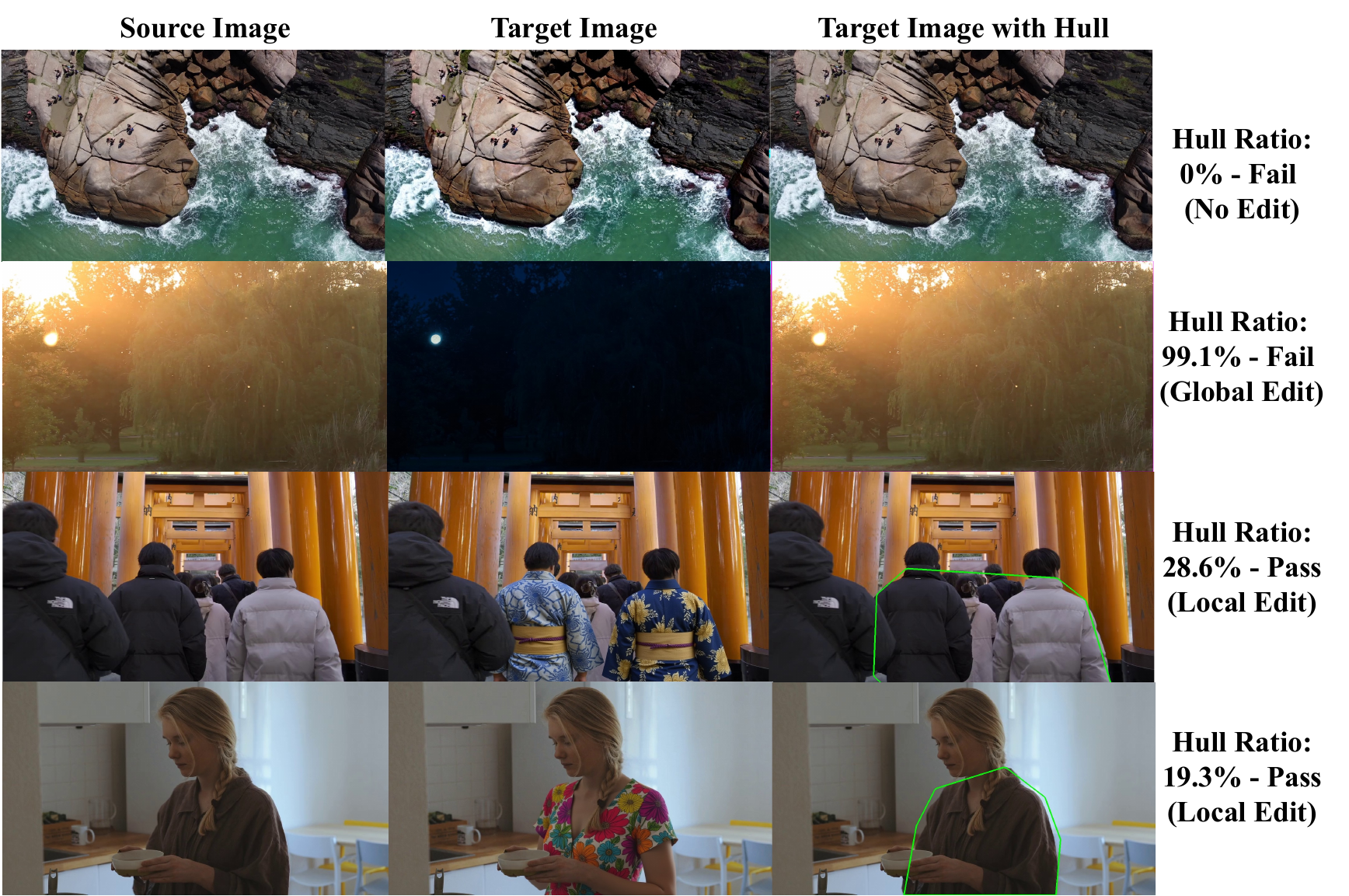}
    \vspace{-0.4cm}
    \caption{Visualization of the data filtering strategy for the Spatial Layer. We calculate the area ratio of the convex hull covering the changed regions between the source and target images. The pipeline automatically rejects failed edits with no significant changes (Row 1) or excessive global changes (Row 2), retaining only high-quality local editing pairs (Rows 3 \& 4) for training.}
    \vspace{-0.4cm}
    \label{fig:appen_mask}
\end{figure}

To derive robust masks $\mathbf{M}$ from these generated pairs, we implement a two-stage difference extraction pipeline. First, a pre-screening step is performed using standard pixel-wise differences. We evaluate the edit magnitude across multiple threshold parameters, filtering out samples where the changing hull area has ratio outside the range $[0.001, 0.75]$ to eliminate insignificant or excessive global edits, as illustrated in \cref{fig:appen_mask}. For the retained pairs, the final mask is generated in the CIELAB color space. We compute the Euclidean distance between the aligned Lab vectors to capture perceptual photometric differences, offering superior robustness to compression artifacts compared to standard RGB. To prevent mask fragmentation, we compute the Convex Hull of the detected change regions, ensuring the mask covers the entire semantic object. Finally, this hull is refined using rolling-circle smoothing to simulate the continuous topology of natural human brushstrokes.

To further augment the model's capability for object removal, we follow \citet{jiang2025smarteraser} to create a synthetic dataset. This involves randomly extracting SAM-based foregrounds and pasting them onto arbitrary backgrounds, treating the pre-paste image as the ground truth target.

\section{User Study for Content Layer}
Standard quantitative metrics often fail to capture perceptual nuances in image composition, such as lighting integration and occlusion handling. To evaluate the human-perceived quality of our method, we conducted a user preference study focusing on the content layer ($L_{fg}$).

\noindent\textbf{Setup and Methodology.} We recruited 30 participants to evaluate 10 editing scenarios sampled from our test set. For each scenario, participants compared results from six methods: Nano Banana \cite{nano_banana}, Insert Anything \cite{song2025insert_anything}, FLUX Kontext \cite{batifol2025kontext}, Kontext + Put It Here LoRA \cite{putitherev4}, Qwen-Image-Edit \cite{wu2025qwen-image}, and ours. Participants selected the single ``best'' result based on foreground preservation, visual coherence, and interaction plausibility.

\begin{figure}[h]
    \centering
    \includegraphics[width=0.9\linewidth]{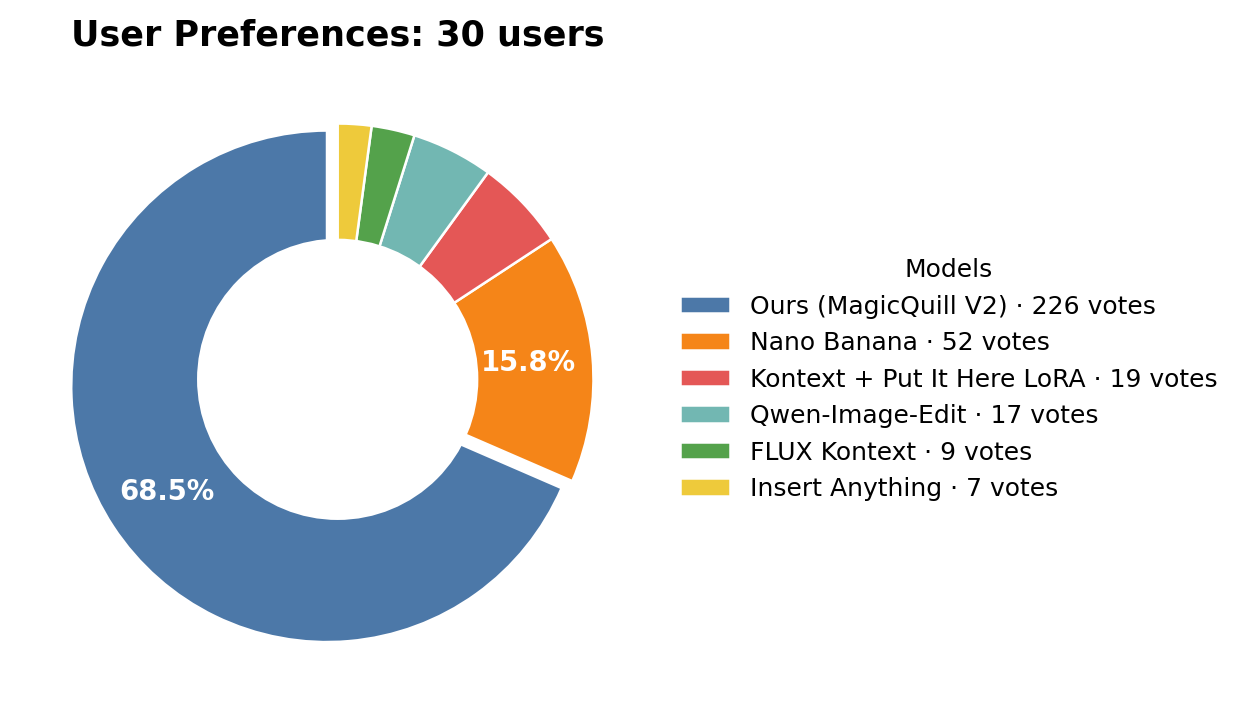}
    \vspace{-0.2cm}
    \caption{\textbf{User Preference Distribution.} MagicQuill V2 (Ours) dominates with \textbf{68.5\%} of the votes, significantly outperforming the strongest baseline, Nano Banana (15.8\%).}
    \label{fig:appen_user_study}
    \vspace{-0.4cm}
\end{figure}

\noindent\textbf{Results.} As shown in \cref{fig:appen_user_study}, \textbf{MagicQuill V2 achieved a dominant preference rate of 68.5\%} (226 votes), significantly outperforming the second-best model, Nano Banana (15.8\%, 52 votes). Qualitative feedback indicates that our method consistently delivers results that are physically grounded and harmoniously integrated.

\section{Limitation}
Despite the significant advancements in precise controllability, MagicQuill V2 has several limitations that present avenues for future research.

\noindent\textbf{Inference Latency.} As a diffusion transformer-based model utilizing a heavy backbone (12B) and multiple control adapters, our system prioritizes generation quality over speed. A single edit requires approximately 30-45 seconds on a high-end H20 GPU. This latency creates friction in the ``interactive'' workflow, as users must wait for feedback. Future work could explore consistency distillation \cite{luo2023lcm,luo2023lcmlora} or quantization techniques \cite{li2024svdquant} to achieve near real-time performance for interaction.

\noindent\textbf{Conflict between Modalities.} While our layered architecture enables flexible composition, the freedom to stack multiple control layers introduces the potential for contradictory inputs. Conflicts can arise not only between semantic text and visual cues (e.g., a prompt describing a ``circle'' while the structural layer dictates a ``square'') but also between distinct visual layers. For instance, a user might provide a detailed structural layer (edge map) that outlines complex geometry, while simultaneously applying a color layer that contradicts those boundaries. Currently, the model attempts to reconcile these disjointed signals based on learned priors and the per-layer control strength $\sigma$. However, when cues are heavily conflicting, this implicit resolution can result in artifacts, where the model fails to satisfy mutually exclusive constraints.

\section{Open Source Commitment}
To facilitate reproducibility and future research, we are committed to fully open-sourcing our project. We will publicly release the complete codebase, including training scripts, model checkpoints, and the interactive system, including the user interface.
\newpage


\end{document}